\documentclass[conference]{ieeeconf}
\IEEEoverridecommandlockouts
\usepackage{cite}
\usepackage{amsmath,amssymb,amsfonts}
\usepackage[ruled, vlined, linesnumbered]{algorithm2e}
\usepackage{graphicx}
\usepackage{textcomp}
\usepackage{xcolor}
\usepackage{tabularx}
\usepackage{listings}
\usepackage{comment}
\graphicspath{{figures/}}
\usepackage{hyperref}

\newcolumntype{M}[1]{>{\centering\arraybackslash}m{#1}}

\usepackage{etoolbox}
\providetoggle{robotresults}
\settoggle{robotresults}{true} 

\providetoggle{validation}
\settoggle{validation}{false}

\providetoggle{playful_explanation}
\settoggle{playful_explanation}{false}

\def\BibTeX{{\rm B\kern-.05em{\sc i\kern-.025em b}\kern-.08em
    T\kern-.1667em\lower.7ex\hbox{E}\kern-.125emX}}

\title{Learning Sensory-Motor Associations from Demonstration}    

\author{\authorblockN{Vincent Berenz$^{1}$, Ahmed Bjelic$^{2}$, Lahiru Herath$^{1,3}$, Jim Mainprice$^{1,2}$}
 \vspace{0.1cm}
 \authorblockA{\tt{
 \small{$^{1}$name.lastname@tuebingen.mpg.de}, 
 \small{$^{2}$st147964@stud.uni-stuttgart.de}
 }}
 \authorblockA{$^1$Autonomous Motion Department, Max Planck Institute for Intelligent Systems ; T{\"u}bingen, Germany}
 \authorblockA{$^2$Humans to Robots Motion Research Group ; HRM ; MLR ; Stuttgart University ; Stuttgart, Germany.}
 \authorblockA{$^3$Department of Control Systems Engineering ; University of Siegen ; Siegen, Germany.}
 \vspace{-0.5cm}
 \thanks{This research was supported in part by the System Mensch alliance and the Max-Planck-Society. Any opinions, findings, and conclusions or recommendations expressed in this material are those of the author(s) and do not necessarily reflect the views of the funding organizations.}
 }

\begin{document}

\maketitle

\begin{abstract}


We propose a method which generates reactive robot behavior learned from human demonstration.
In order to do so, we use the Playful programming language which is based on the reactive programming paradigm. 
This allows us to represent the learned behavior
as a set of associations between sensor and motor primitives in a human readable script.
Distinguishing between sensor and motor primitives introduces a supplementary level of granularity
and more importantly enforces feedback, increasing adaptability and robustness.
As the experimental section shows, useful behaviors may be learned
from a single demonstration covering a very limited portion of the task space.

\end{abstract}

\section{Introduction}

Behavior orchestration is the offline
setup of activation rules of software primitives
so that a robot autonomously performs 
a desired behavior online.
In this paper, we propose a new method to generate 
behavior orchestration from human demonstration.
The novelty of the proposed method is that it 
does not only learn the activation rules of primitives:
it also learns how information should be exchanged 
between sensor primitives
and motor primitives, i.e. it learns which sensory-motor
associations should be performed by a robot
to imitate a demonstrated
behavior.


When composing a behavior from primitives, 
most methods encode the rules for transiting from one sub-behavior to the other.
This may result in learning state transitions of a state machines
or in generating a behavior tree in which edges encode the logic,
and module activation relies on traversing the tree during run-time (see section \ref{sec:related_work}).
While these methods use one type of primitive, we describe in this paper a method which distinguishes between sensor and motor primitives. 
The system does not only learn the activation rules of primitives,
but also the associations that should be performed between sensor and motor primitives.

For example, a ``look at" motor primitive could be associated to a human detection sensor primitive, creating the sensor-motor coupling
required for the robot to look at a moving human. The same ``look at" primitive
could be associated with a ball detection sensor primitive, resulting
in the robot looking at a moving ball. The demonstrator may also during demonstration alternate between looking at the ball and looking at her/his human partner. The goal of our approach is to find which sensor-motor associations were performed by the demonstrator,  
and the underlying rules the demonstrator applied to activate
these associations.

Furthermore, when parsing the data collected during a human demonstration, 
the proposed method does not attempt to map it to a sequence of actions
the robot may replicate. Instead, it detects which sensory-motor
associations the demonstrator performed and then encodes
the demonstrated behavior into non-sequential rules of activation of sensor-motor associations
(e.g. ``look at the ball when close to it"). 

Finally, it organizes the behavior hierarchically, i.e., in a dynamic behavior tree.

The proposed approach has two particularities:

\begin{itemize}
    \item   It relieves the algorithm of generalizing observed task-space trajectories. For example, if the demonstrator directed herself toward an object, the algorithm will only learn that the sensory-motor loop between the detection of this object and the wheels of the robot is of interest. During run-time, the trajectory resulting from the activation of primitives when the robot directs itself toward the object will be different than the one observed during demonstration. It will nevertheless matches the intent of directing itself toward the object.  
    \item The method generates behaviors that do not encode explicit goals to achieve, e.g. ``first grasp the bottle then fill the glass''. Instead, sequences are implicitly encoded, e.g. ``if no bottle in hand, activate primitives to grasp the bottle; if bottle in hand, activate the primitives to pour in the glass''. This example shows the generalization benefits: if the bottle is removed from the hand of the robot, an explicitly encoded sequence would be interrupted and require re-planning. The applied rule-based approach does not suffer from such difficulty. 
\end{itemize}

The output of the proposed algorithm is an executable script
that can be interpreted by Playful. Playful is an interpreter for behavior orchestration using the reactive programming paradigm. 
A Playful script consists of a series of declarative statements. Each statement relates a target (the output of a sensor primitive) to both a motor primitive (such as ``look at'') and an activation rule (such as ``when close to the robot''). The order of the statements in the script does not matter. Instead, at run-time, all rules of activation are continuously evaluated,  and the motor primitives are activated or deactivated accordingly. Furthermore, the scripting language allows to group statements hierarchically to form new reusable higher-level primitives.  This results in behavior trees, in which the activation status of branches are monitored online.
Playful is further described in section \ref{sec:framework}, but from this short overview, it can be understood that a Playful script encodes sensory-motor associations as presented above. It can be noted, that if in this paper the algorithm outputs a Playful script, the proposed approach is general and could be applied for usage with other reactive programming interpreters.

Because it distinguishes between sensor and motor primitives, the proposed method introduces a supplementary level of granularity: primitives are no longer the building blocks available to the algorithm for imitating the human demonstration. Instead, sensor-motor associations are. If the library of primitives has $n$ sensor primitives and $m$ motor primitives, then there are $n \times m$ sensor-motor associations available to the learning algorithm.
Encoding the behavior using a high number of possible associations allows for the learning of rich interactive tasks. Furthermore, the sensor-motor associations implement feedbacks that by nature increase adaptability and robustness. Consequently, the demonstrated behavior is encoded in a fashion that grasps the intent and motion of the demonstrator beyond the sequence of high-level actions she or he performed.  

In our experiments (section \ref{sec:experiments}), we consider an example where a
human attempts to attract the attention of another human passing by,
combining head and arm movements with walking. 
The scripts, which are reported in the paper, are produced from a single demonstration
and remain compact while capturing the essence of the performed demonstration.

\section{Related work}
\label{sec:related_work}


\subsection{Learning from Demonstration}
\label{sec:LfD}

Learning from human teachers \cite{Chernova:2014cja},
has been a subject of study in the robotics community for
decades. It raises multiple questions, including
how humans should convey the demonstration and
how to design learning algorithms \cite{Osa:2018jc}.
LfD algorithms are classified
as 1) Behavior Cloning (BC) or Inverse Reinforcement Learning (IRL),
depending on whether the policy is encoded directly or through a reward function,
and as 2) Task or Motion Control learning, depending on the nature of the state-action space.
In this work we consider high-level LfD by encoding the policy directly
into an executable Playful script, which describes a dynamic Behavior Tree as presented in Section \ref{sec:behavior_trees}.


In high-level BC, the focus is to learn multi-step policies
that combine sequences of primitive actions.
Generally the corresponding approaches are broken into three parts:
1) segmentation of the demonstrations by identifying
substructures, 2) learning the primitive actions and 3) learning
the transitions between primitives.
Typically the segmentation is performed using Hidden Markov Models (HMMs).
For instance in \cite{Niekum:2012learning}, demonstrations are segmented using a beta-process HMM,
and the primitives are learned using Dynamic Movement Primitives \cite{Schaal:2006dynamic}.
In this paper, we rather consider fixed pre-programmed primitives,
divided between sensor and motor primitives.
Instead of considering a sequence of active action primitives, we consider
a sequence active sensor-motor couplings, where multiple couplings can be active simultaneously.
This approach leads to more reactive behaviors and the specification is more modular.

The sequencing of primitive actions can also be learned
hierarchically by encoding them in a Hierarchical Task Network (HTNs).
In recent work \cite{Mohseni:2015interactive}, authors have
described the encoding of the task structure by having a demonstrator
narrating the task. This approach however does not allow
for the activation of multiple primitives simultaneously.

\subsection{Behavior Trees}
\label{sec:behavior_trees}

Behavior Trees (BTs) yield modularity and reactivity,
by organizing the switching structure of hybrid dynamical systems.
BTs have initially emerged in the computer gaming field
as a replacement to Finite State Machines (FSM) for encoding the behavior of non players characters.
In \cite{Colledanchise:2017tro}, Colledanchise and {\"O}rgen have formalized
BTs and shown that they generalize the sequential behavior composition,
the subsumption architecture and the decision trees.

In a sense, BTs are conceptually close to HTNs,
which have been used for planning for three decades \cite{Erol:1994wh}.
However BTs implement dynamic behaviors which are closely related
to supervision in robotic systems.
Thus they have been applied
for high-level supervision of manipulation tasks  in
\cite{Bagnell:12, Scheper:2014uv}, and an algorithm has been recently proposed
to synthesize BTs using high-level planning \cite{Colledanchise:2019fe},
connecting high and low-level reasoning.

In this work we learn Playful scripts \cite{Berenz:2018eq}, which differs from the standard BT framework \cite{Colledanchise:2017uk} by:
\begin{enumerate}
    \item dividing action primitives in sensor and motor primitives
    \item encoding the associations between these sensor and motor primitives
    \item evaluating the BT at a fixed frequency,
leading to activation and/or deactivation of its branches, resulting in higher behavioral reactivity. This differs from the previous aforementioned approaches in which logic is encoded in edges, and module activation relies on
traversing the tree during operation.
\end{enumerate}



\subsection{Learning Behavior Trees from Demonstration}

Despite the various works on learning from human demonstration,
to the best of our knowledge only few works have been carried in learning hybrid controllers from demonstration.
The earliest work was performed using a subsumption architecture in \cite{Kasper:01}.
However, this work focuses on navigation behaviors and does not learn a hierarchical grouping of primitives.

The closest to our approach are \cite{Sagredo:2017trained, Colledanchise:2017tro, Colledanchise:2017uk, French:2019ve}.
In these works a BT is learned from demonstration by exploiting
the equivalence between BTs and decision trees \cite{Colledanchise:2017tro}.
In \cite{Sagredo:2017trained}, the C4.5 algorithm is used to first learn a decision tree.
The tree is first flattened, and then simplified using a greedy algorithm.
In \cite{French:2019ve}, the authors extend this work by taking into account any
logic using the UC Berkeley Expresso algorithm \cite{Brayton:1984logic}.
Our work shares similarities with these works, however because we separate
sensor and motor primitives and produce a Playful script, we use a dedicated algorithm
to convert the decision tree.

\section{Learning a Playful Script}
\label{sec:framework}

\subsubsection{Playful Framework}
\label{sec:playful_example}


Playful is a software for the orchestration of robot behaviors \cite{Berenz:2018eq}.
It allows developers to compose applications via a list of declarative statements such as:

\begin{lstlisting}[float=h!,
basicstyle=\fontsize{7.5}{7.5}\ttfamily,breaklines=true,
caption={Playful grasping example},
captionpos=b,
label={alg:example},
keepspaces=true,
keywordstyle=\bfseries\color{blue},
morekeywords={targeting, whenever, priority},
escapechar=\&]
ball_detection | out=ball
head_search, priority of 1
targeting ball: look_at, whenever seen, priority of 2
targeting ball: grasp, whenever close 
\end{lstlisting}

This playful script has a robot running ball detection (line 1), searching for a ball (line 2), 
looking at the one it detected (line 3) and attempt grasp when the ball is close to the robot (line 4). Sensory-motor associations between the ball detection sensor primitive and the motor primitives (\textit{look\_at} and \textit{grasp}) are specified via the \textit{targeting} keyword.

Note that the motor primitives are not associated directly to the sensor primitive, rather
they are associated with the corresponding \textit{target}. A target is a discretized sensory information item which data is continuously updated by the sensor primitive. 
A target is used as an association proxy between a sensor and a motor primitive, e.g \textit{look\_at} is associated to \textit{ball\_detection} via the \textit{ball} target. 
In the rest of the paper, we use indifferently target-motor association and sensor-motor association.

Playful is based on reactive programming, i.e. the order of statements is of no importance,
and no sequence of action is explicitly programmed. The three statements are continuously evaluated,
and their \textit{primitives} (here \textit{head\_search}, \textit{look\_at} and \textit{grasp})
are activated when their
\textit{evaluations} (here, \textit{seen} and \textit{close}) return true.
 A \textit{priority} is used to solve conflicts, for example \textit{head\_search} and 
 \textit{look\_at} can not be active simultaneously because they share the resource \textit{head}.
Here, the higher priority of 2 for \textit{look\_at} takes precedence.

Evaluations, i.e. the activation rules, correspond to python code, e.g., \textit{close} returns true if the
distance between the robot and the ball is below a given threshold.

Playful scripts encode \textit{behavior trees} in which nodes may
be \emph{primitives} that correspond to python code
communicating with the robot middleware;
but may also correspond to other Playful scripts.
For example, in the grasping scenario of Listing \ref{alg:example},
\textit{grasp} could be a Playful script consisting
of two statements, e.g. move\_arm and move\_hand.

Thus, the status of evaluations and priorities results in the activation and
deactivation of branches at run-time, shaping the behavior of the robot.
For a complete formalized description of Playful, we invite readers to refer to \cite{Berenz:2018eq}.

\vspace{.3cm}
\subsubsection{Problem Statement}


Given a list of Playful sensor and motor primitives,
we propose the automated generation of a Playful script
based on data collected during human demonstration.


At the start of each learning from demonstration phase,
the learner is given a set of observations (i.e., LfD dataset):
\begin{equation}
\mathcal{D} = \{ x_t \}_{t=1:T} \; ,
\end{equation}

\noindent
where $t$ is a time index and $x_t$
encodes positions and velocities
of the humans and objects in the scene.

\paragraph{Flat tree}
The state dynamics $x_{t+1} = f_\theta(x_t, z_t)$ resulting from the execution of a Playful script, can be expressed by the following function:

\begin{equation}
\label{eq:flat_tree}
 f^{(1)}_\theta(x_t, z_t) =
\begin{cases}
a_1(x_t, z_t) & \text{if} \; e_{\theta_1}(x_t) > 0\\
\vdots \\
 a_n(x_t, z_t) & \text{if}\;  e_{\theta_n}(x_t) > 0\\
\end{cases},
\end{equation}

\noindent
where $z_t$ is the Playful memory at time index $t$,
$a_i$ is the result of the activation of the $i$th leaf node
(i.e., sensory-motor primitive association in each branch), and
$e_{\theta_i}: X \to \{0, 1\}$ are the binary conditional-evaluation maps
parametrized by some vector $\theta_i$.

We will subsequently call this representation, the \textit{flat tree}, 
i.e. the raw list of sensory-motor association without any hierarchical grouping.
Note, that we have relaxed the definition of $f_\theta$
by leaving out the priority mechanism.

\paragraph{Resources}
The simplified definition of a Playful behavior in Equation \ref{eq:flat_tree}
only holds for the case where evaluations $e_{\theta_i}$ are disjoint.
This is not the case in practice, however when the leaf nodes $a_i$ control a given \textit{resource},
e.g., specific DoFs such as the arm, the base or the head,
the actions are limited to separate dimensions of the range space of $ f_\theta$.
In this case, the \textit{flat tree} can be expressed as follows:
\begin{equation}
\label{eq:function_playful}
f^{(2)} _\theta(x_t, z_t)= \sum_i e_{\theta_i}(x_t)  \; a_i(x_t, z_t).
\end{equation}

\paragraph{Learning Loss}
Given the priority simplification, learning a Playful script
amounts to finding for each resource the parameters $\theta$ that minimize
the classic LfD mean squared-error loss:
\begin{equation}
\label{eq:problem}
L(\theta) = \frac{1}{T}\sum_t \| x_{t+1} -f^{(2)}_\theta(x_t, z_t) \|^2 \; \forall x_t \in \mathcal{D}.
\end{equation}

The overall behavior is the union of all \textit{flat trees} found for each resource,
and a hierarchical grouping of nodes can be performed based on activation pattern similarities of the leaf nodes $a_i$.
Note that there is always an equivalent flat tree
corresponding to a hierarchical tree. 
The composition of a tree with several layers is beneficial to simplify
the output script and provide reusable behaviors.

\begin{algorithm*}[t!]
  {
  \label{alg:LRPD}
    \textbf{input} : $\mathcal{D} = \{ x_t \}_{t=1:T}$ sequence of observed states  \\
    \Begin{
    	$\mathcal{O} \leftarrow$ Compute target-primitive emission densities $p(o^i_t | w^i_t)$ for all time steps in $\mathcal{D}$\;
    	\For{ All resources }{
		$\mathcal{W} \leftarrow$ Find target-primitive activations using Viterbi on the HMM with emissions densities $\mathcal{O}$ \;
		$\Theta \leftarrow$ Train evaluations for active behaviors as binary classifiers over $(x_t, w_{t}^i) \in \mathcal{D} \times  \{0, 1\}  $ \;
		$\mathcal{T} \leftarrow$ Convert activations and evaluations $\mathcal{W}$,  $\Theta$ to \textit{flat tree} \;
		$\mathcal{T} _{flat} \leftarrow \mathcal{T} _{flat}\cup \mathcal{T}$ \;
		}
		$\mathcal{T}_{H} \leftarrow$ Find hierarchies in $\mathcal{T}_{flat}$ \;
		script.pf  $\leftarrow$ Convert $\mathcal{T}_H$ to Playful program \;
	}
    \caption{Learning Reactive Programs from Demonstrations (LRPD) }
  } 
\end{algorithm*}

\section{Learning Reactive Programs from Demonstrations}
\label{sec:algorithm}

Algorithm \ref{alg:LRPD} sketches our LfD method.
Intuitively, the core of the algorithm can be viewed
as being separated in the 3 flowing phases:
\begin{enumerate}
\item detect the most likely sequence of target-primitive activations 
		  based on a probabilistic inverse model,
\item train evaluations based on state-activation pairs (i.e, input-output) using supervised learning,
\item factorize the \textit{flat tree} and generate the program.
\end{enumerate}


\subsection{Target-primitive Activation Detection}
\label{sec:target_primitive_activation_detection}

The first step of LRPD (i.e., line 3 and 5 of Algorithm \ref{alg:LRPD}), 
is to detect which target-primitive associations are active at each time step $t$
of our dataset $\mathcal{D}$.
To do this we map the state space of a Hidden Markov Model (HMM)
to activation patterns \mbox{$w^i : t \mapsto \{0, 1\}$} at each time step $t$
of target-primitive associations $i$
(e.g., ball-look\_at in the example of Section \ref{sec:playful_example}). 

We set the uniform transition probabilities and
model the observation probability at time $t$
of one such association $o^i_t$  by the following
Gaussian emission densities:
\begin{equation}
p(o^i_t | w^i_t) \propto \exp \bigg(-\frac{| \phi_i(x^i_t) -  \phi_i(x_t) |^2}{2\sigma_i} \bigg)  
\end{equation}

\noindent
where $x^i_t$ is the state we would observe if the $i$th
target-primitive was active. $\phi_i$ is some non-linear
function of the state defined for the target-primitive $i$,
(e.g., angular velocity when turning, heading direction when moving forward).
$\sigma_i$ is a standard deviation term,
which is parameter of the algorithm.
We make use of
the Viterbi algorithm to recover the most likely sequence of activation $\mathcal{W}$
for all the combinations of targets and primitives.

\subsection{Binary Classifiers}
\label{sec:training}

Once the data points $x_t$ are labeled with 
target-primitive activation $w_t^i$,
evaluation functions $e_{\theta_i}$ can be trained to activate
the target-primitive for each time index $t$.
See line 6 of Algorithm \ref{alg:LRPD}. 
In general, any binary classification method could be used
(e.g., support vector machine, neural network).
However, in our experiments, we define a non-linear feature $\phi_d : X \to \mathbb{R}$,
which models the human-robot interaction distance, and simply identify intervals of its co-domain.

\subsection{Factorization of the Tree}

The last step of the algorithm consists of unifying
the flat trees $\mathcal{T}_{flat}$ defined for each resource and grouping the nodes in hierarchies.

Grouping of nodes is performed by measuring similarity of their conditional evaluation functions $e_{\theta}$.
Similarity between $e_{\theta}$ is difficult to define in the general case.
In this work we simply compare the subset of the domain of $e_{\theta}$ where the
evaluation is true. This matching is performed up to some error $\Delta$,
which trades off complexity for fidelity.

\section{Experimental Results}
\label{sec:experiments}

We used the proposed approach to generate scripts that have a mobile humanoid robot (Softbank Robotics Pepper \cite{Pandey:2018}) applying strategies to attract passing visitors to a stand.
Each demonstration was the result of capturing two humans
(i.e., a demonstrator to imitate, and a visitor).

In this section, we provide implementation details, and show how the learned robot behaviors
capture successfully the different strategies applied by the demonstrators.

\subsection{Experimental setup}

\begin{figure*}[t]
\centering
\includegraphics[scale=0.47]{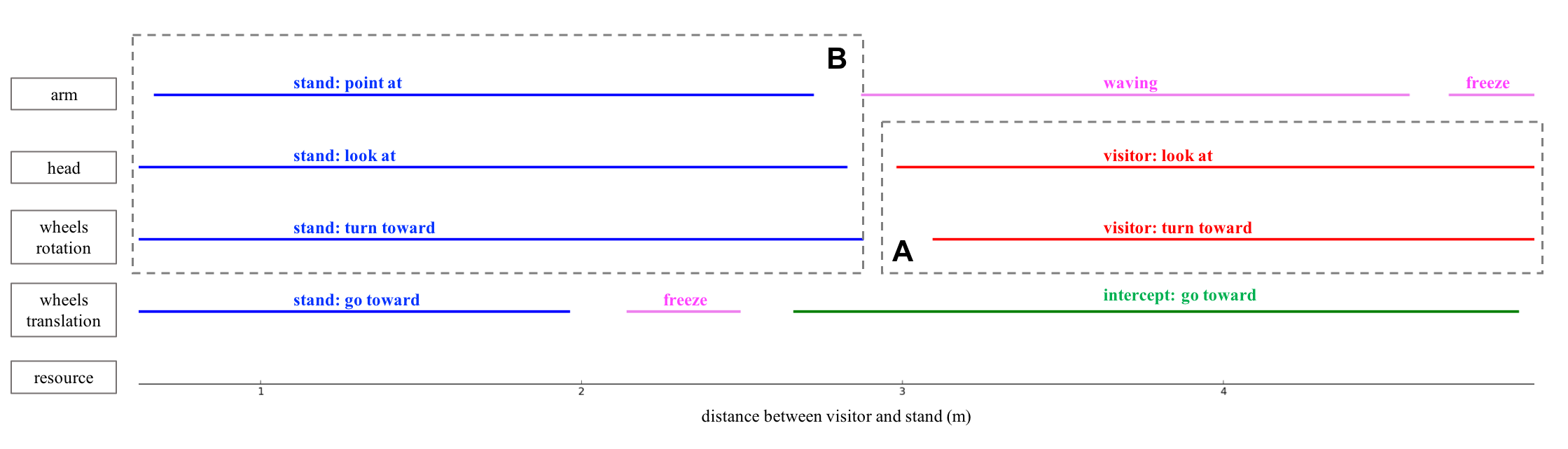}
\caption{For each resource, the most likely sequence of activation of target-primitive associations for demonstrator 1 as function of
the distance to the stand.
Dotted squares present the grouping of target-primitives applied with a distance precision of 30 cm, as a pre-process for the generation of the algorithm presented in listing \ref{listing:subject1}.
\textbf{A} and \textbf{B} refer to the labeling of the resulting branches in the Playful script.
From right to left (decreasing distance between the visitor and the stand), it can be seen that the demonstrator first walked to the front of the stand (green color) while turning toward and looking at the visitor (red color); and waving the hand (violet). As the visitor approached, the demonstrator started to redirect himself toward the stand while pointing at it (blue color). }
\label{fig:states}
\end{figure*}

\subsubsection{Robot}

We made use of a Pepper robot from Softbank Robotics,
which is 1.2 meters in height and offers a total of 20 DoFs, 17 for the body and 3 for the base
(see Figure \ref{fig:demos}).
The base is omnidirectional and allows for holonomic navigation.
Pepper is equipped with an IMU, which coupled with the wheel's encoders, provides odometry.
The odometry was used to update the position of the stand relative
to the robot during all the experiments.

In this study, we used two of the three cameras available on the robot.
The first is an RGB camera with a native resolution of 640*480 positioned in the forehead. 
The second is an ASUS Xtion3D sensor located in one of its eyes.
We used these cameras in combination with OpenPose \cite{cao:2018} for 3D human detection.
The other sensors of the robot (sonars, infrared sensors, touch sensors and bumpers) were not used.

\begin{table}[htb]
\caption{Motor Primitive descriptions}
\label{table:primitives}
\footnotesize
\begin{center}
    \begin{tabular}{ |l|l|}

    \hline
    \textbf{Primitive} & \textbf{Description}\\ \hline \hline

    \multicolumn{2}{|l|}{resource: \textbf{wheels rotation}} \\ \hline
    turn\_toward &  rotates body to face the target \\ \hline
    turn\_stop & angular velocity of the robot set to 0 \\ \hline \hline

    \multicolumn{2}{|l|}{resource: \textbf{wheels translation}} \\ \hline
    go\_toward & moves toward the target, keeping a safety distance \\ \hline
    go\_stop & linear velocity set to 0 \\ \hline \hline

    \multicolumn{2}{|l|}{resource: \textbf{head}} \\ \hline
    look\_at & moves to keep the target object in the center of FoV \\ \hline \hline

    \multicolumn{2}{|l|}{resource: \textbf{arm}} \\ \hline
    point\_toward & moves left arm to point toward the target  \\ \hline
    waving &  waves the left arm   \\ \hline
    arm\_freeze & stops moving the left arm \\ \hline

    \hline

    \end{tabular}
 \end{center}
\end{table}

\subsubsection{Available Playful Primitives}

Previously and independently of this study, we implemented a set of basic Playful primitives for Pepper,
which are listed in Table \ref{table:primitives}.
Each primitive is associated with a resource, i.e. the robotic joint required for its activation. Primitives associated to the same resource may not activate at the same time. Because of the holonomic capabilities of the mobile base, wheels rotation and wheels translation refers to two separated resources.

Out of the eight primitives, four of them (\textit{turn\_toward}, \textit{go\_toward}, \textit{look\_at} and \textit{point\_toward}) have to be associated to a target for activation. In this study, three targets are set to be tracked by the Playful backend: the visitor (human detected using the cameras), the stand, and a virtual fixed position in front of it (tracked using odometry). This results in a total number of 16 possible primitives or target-primitives association that may be activated.

For the application of the algorithm presented in Section \ref{sec:algorithm},
an inverse model $\phi_i$ for each primitive has been defined
to implement observation probabilities $p(o_t^i| w_t^i)$.
These models express the likelihood of observing a certain change in
the state of the world given that the motor-coupling $w_t^i$ is active
(e.g., for look at we consider the angle between target and the demonstrator field of view).
All $\phi_i$ were normalized before training.

\subsubsection{Human demonstration}

Motion capture was performed using a Vicon tracking device.
Each dataset $\mathcal{D}$
captured
the (fixed) position of the stand,
the position of the visitor, the pose of the body of the demonstrator,
the pose of the head of the demonstrator, and motion of the left hand of the demonstrator
using a stick equipped with reflective markers.

Two participants (i.e, demonstrators) were recruited and given the instructions
to try to attract a passing visitor to a stand.
We required that he/she should 1) not touch persons or objects, 2) only use the left arm, and keep the right arm inactive and 3) apply common courtesy (e.g. not trying to attract to the stand by using physical constrain).
Apart from these limitations, subjects were instructed to move freely in the experimental space. 
The subjects were not informed of the objective of the experiment.

The experimental space consisted of a corridor of 5 meters in length and 3 meters of width.
The stand, a small pulpit, was located at the end and side of the corridor. 
The demonstrator starts at the stand, and waits for the visitor to enter the corridor.
The visitor then enters the corridor, walking straight, before diverging toward
the stand when at around 2 meters from it.
Data recording stops when the visitor reaches the stand.

The demonstrations are shown in the associated video\footnote{\url{https://youtu.be/-8pUC0rN-fQ}}.
It can be seen that the two demonstrators apply different strategies.
 Demonstrator 1 goes in front of the stand while actively pointing towards it.
 Demonstrator 2 goes toward the visitor before backing toward the stand.

\subsubsection{Algorithm run}
\label{sec:algo_run}

\def\playful_fontsize{7.}

\begin{lstlisting}[float=t,basicstyle=\fontsize{\playful_fontsize}{\playful_fontsize}\ttfamily,breaklines=true,caption={Script generated from the demonstration of demonstator 1. \textit{d} refers to the distance between the visitor and the stand.},captionpos=b,label={listing:subject2},keepspaces=true,keywordstyle=\bfseries\color{blue},morekeywords={targeting,whenever,A,B},escapechar=\&]

wheels_translation_freeze, whenever d in [2.1,2.5]
targeting stand: go_toward, whenever d in [0.6,2.0]
targeting visitor: A, whenever d > 2.7
targeting intercept: go_toward, whenever d > 2.7
arm_freeze, whenever d > 4.7
waving, whenever d in [2.9,4.6]
targeting stand: B, whenever d < 2.7

A:
    turning_toward
    looking_at

B:
    A
    pointing_at

\end{lstlisting}

\begin{lstlisting}[float=t,basicstyle=\fontsize{\playful_fontsize}{\playful_fontsize}\ttfamily,breaklines=true,caption={Script generated from the demonstration of demonstrator 2. },captionpos=b,label={listing:subject1},keepspaces=true,keywordstyle=\bfseries\color{blue},morekeywords={targeting,whenever,A,B},escapechar=\&]

wheels_translation_freeze, whenever d > 4.8
targeting visitor: go_toward, whenever d in [3.2,4.6]
targeting stand: go_toward, whenever d in [0.7,3.0]
targeting visitor: A, whenever d < 5.1
arm_freeze, whenever d > 4.2
waving, whenever d in [3.8,4.1]
targeting waving: pointing_at, whenever d in [0.8,3.7]

A:
    turning_toward
    looking_at


\end{lstlisting}

Algorithm \ref{alg:LRPD} was applied, computing the most likely sequence of activation of target-primitive associations. 
These are presented for demonstrator 1 is represented in Figure \ref{fig:states} as a function of decreasing distance between the visitor and the stand. As the visitor approached, the demonstrator started to redirect himself toward the stand while pointing at it. 
From this sequence, overlapping target-primitives are then grouped together using a similarity score implementing a tolerance of 0.3 meter (dotted squares in Figure \ref{fig:states}). These groups, labeled A and B, corresponded to branches in the subsequently generated Playful scripts. Related evaluations $e_{\theta}$ consisted of the boundary in terms of distance between the visitor and the stand observed during activations. 

\subsection{Analysis of the learned scripts}

The resulting Playful script is shown in Listing \ref{listing:subject2}. It encodes this demonstrated behavior. For example, when the visitor is far (distance superior to 5.1), the branch A activates targeting the visitor, i.e. the robot is commanded to turn toward the visitor while looking at it (line 3). Similarly, when the visitor gets closer (distance below 2.7), the branch B activates, commanding the robot to look at the stand while turning and pointing towards it (line 7).

The same procedure was applied for demonstrator 2, resulting in Listing \ref{listing:subject1}.
This script reflects the different strategy applied by the second demonstrator,
for example first approaching the visitor (line 4) before backing toward the stand (line 3).

\subsubsection{Implementation and result}

\begin{figure}[t!]
\centering
\includegraphics[width=.9\linewidth]{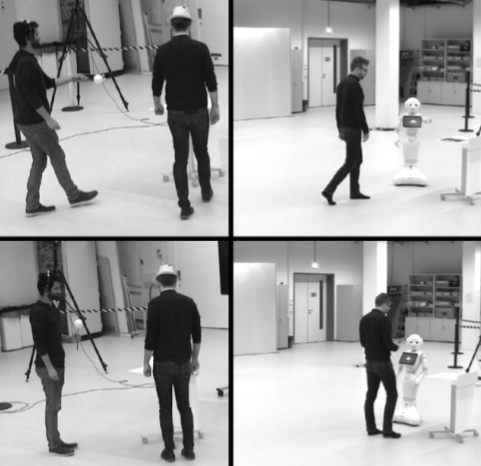}
\caption{Pepper robot (right) attempting to attract a visitor to a stand by running a script generated from human demonstration (left)}
\label{fig:demos}
\end{figure}

To the generated Playful script were added two nodes:
\begin{itemize}
 \item A higher priority security node, that has the robot safely deactivating itself when its battery is low, or its head is touched
 \item A lower priority default behavior node, that has the robot moving randomly the arms and the head, while going back to the stand. This node provides a default simple set of primitives that activate when the visitor is not detected.
\end{itemize}

The generated scripts were not edited in any other way, and directly executed on the robot, as shown in 
Figure \ref{fig:demos}. The resulting behaviors can also be seen on the support video, 
where it can be observed that the robot replicates with fidelity the high level strategy of each demonstrators as described in the previous section (\ref{sec:algo_run}).

\subsection{Analysis of the learned behavior}

\begin{figure}[t!]
\centering
\includegraphics[width=.7\linewidth]{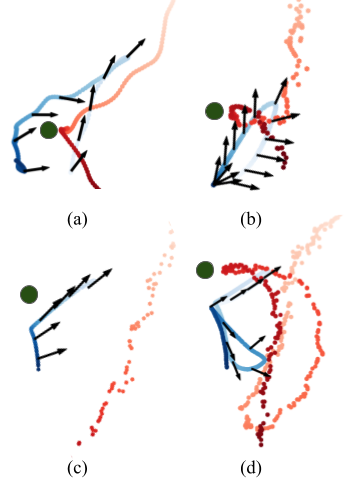}
\caption{(a) Orientation and trajectory of demonstrator 2 (in blue) and of the visitor (in red), as captured by the vicon tracking system. (b), (c) and (d) Orientation and trajectory of the robot (in blue) and of the visitor (in red) as captured by odometry and the xtion camera embedded in the robot. See text in section \ref{sec:experiments} for details}
\label{fig:generalization}
\end{figure}

Figure \ref{fig:generalization}.a displays in blue the demonstration performed by the second user
in the video attachment. The arrows represent his body orientation.
The trajectory of the passing visitor is represented in red.
Changes in intensity (from light to dark) represent passing time. The visitor first goes toward the stand before walking away.
The demonstrator first goes toward the visitor before reverting toward the stand (i.e., green dot),
his body always turning toward the visitor.

Figures \ref{fig:generalization}.b,c,d show three executions of the learned script on the Pepper robot
shown on Figure \ref{fig:demos}. In (\ref{fig:generalization}.b), the visitor has a behavior similar as 
during the demonstration. In (\ref{fig:generalization}.c), the visitor passes by without visiting the stand. In (\ref{fig:generalization}.d), the visitor first pass-by the stand, then comes back to it and finally leaves.

The behavior performed by the robot is robust to the visitor diverging from the demonstrated behavior.
In the three cases, the robot properly either goes toward the visitor or back toward the stand.
In addition, similarly to the demonstrator, the robot keeps his body turned
toward the visitor at all time, which shows how our approach
is able to capture the ``spirit" underlying the task using a single demonstration.

\section{Conclusion}

We propose a new method of learning from demonstration, which encodes the learned behavior as sensory-motor associations organized hierarchically. Because of the feedback it enforces, the method can robustly generate Playful scripts based on a single demonstration. A direct benefit of encoding behavior into Playful script is the liveliness and behavioral reactivity displayed by the robot during runtime.
The described algorithm is first order as it  does not encode rules 
for generating an explicit sequence of actions suitable to reach an explicit goal.

Next we will investigate how this approach
can be used to learn evaluations end-to-end from the raw kinematic states directly and how it
could work alongside another learning system which allows for higher level planning.

\pagebreak






\addtolength{\textheight}{-2cm}   

\vspace{.3cm}
\bibliographystyle{ieeetr}
\bibliography{bibliography,playful}

\end{document}